%% file: main.tex
\documentclass[lettersize,journal]{IEEEtran}

\input{header}

\begin{document}

    \title{
        Probabilistic Mission Design for \\
        Neuro-Symbolic Unmanned Aircraft Systems
    }
   
    \author{
        Simon Kohaut$^{1}$, Benedict Flade$^{2}$, Daniel Ochs$^{1}$, Devendra Singh Dhami$^{3}$, Julian Eggert$^{2}$,  Kristian Kersting$^{1, 4, 5, 6}$
        \thanks{
            $^{1}$ Artificial Intelligence and Machine Learning Lab, \newline\hspace*{1.6em} 
            Department of Computer Science, \newline\hspace*{1.6em}
            TU Darmstadt, 64283 Darmstadt, Germany \newline\hspace*{1.6em}
            {\tt\small firstname.surname@cs.tu-darmstadt.de}%
        }%
        \thanks{
            $^{2}$ Honda Research Institute Europe GmbH, \newline\hspace*{1.6em} 
            Carl-Legien-Str. 30, 63073 Offenbach, Germany \newline\hspace*{1.6em}
            {\tt\small firstname.surname@honda-ri.de}
        }%
        \thanks{
            $^{3}$
            Uncertainty in Artificial Intelligence Group, \newline\hspace*{1.6em}
            Department of Mathematics and Computer Science, \newline\hspace*{1.6em}
            TU Eindhoven, 5600 MB Eindhoven, Netherlands%
        }
        \thanks{
            $^{4}$ Hessian AI
        }%
        \thanks{
            $^{5}$ Centre for Cognitive Science
        }%
        \thanks{
            $^{6}$ German Center for Artificial Intelligence (DFKI)
        }%
    }

    \backgroundsetup{
      contents={
          \begin{tikzpicture}
            \node at (current page.center) [align=center] {\textcopyright 2025 IEEE. Personal use of this material is permitted. \\ Permission from IEEE must be obtained for all other uses, in any current or future media, including reprinting/republishing this material for advertising or promotional purposes, \\ creating new collective works, for resale or redistribution to servers or lists, or reuse of any copyrighted component of this work in other works.
             \\ DOI:10.1109/TITS.2025.3609835};
        \end{tikzpicture}},
      placement=bottom,
      scale=0.6,
      vshift=20
    }
        
    
    
    \maketitle

    \begin{abstract}
        \input{content/0_abstract}
    \end{abstract}
    
    \input{content/1_introduction}
    \input{content/2_related_work}
    \input{content/3_methods}

    \input{content/4_experiments}
    \input{content/5_conclusion}
    \input{content/6_acknowledgment}

    \bibliographystyle{IEEEtran}
    \bibliography{references.bib}
    
    \begin{IEEEbiography}[{\includegraphics[width=0.9in,keepaspectratio]{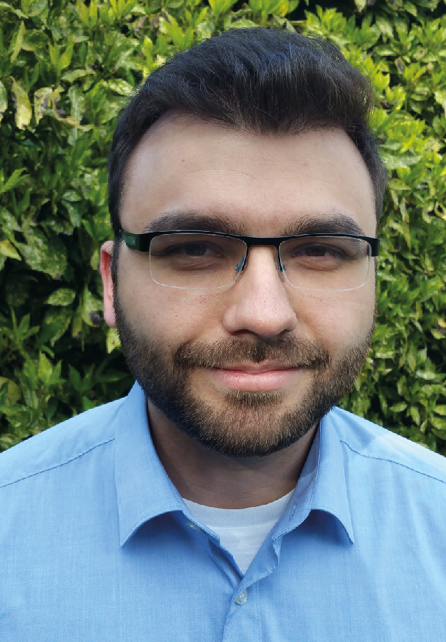}}]
    {Simon Kohaut} graduated from the Technical University of Darmstadt, Germany, with a bachelor's degree in Computer Science and a master's degree in Autonomous Systems.
    Since 2022, he is pursuing his Ph.D. with the Artificial Intelligence and Machine Learning Lab in collaboration with the Honda Research Institute.
    His fields of interest are Neuro-Symbolic systems with a focus on probabilistic logic and its application in navigation under time and safety constraints.
    \end{IEEEbiography}%
    \begin{IEEEbiography}[{\includegraphics[width=0.9in,keepaspectratio]{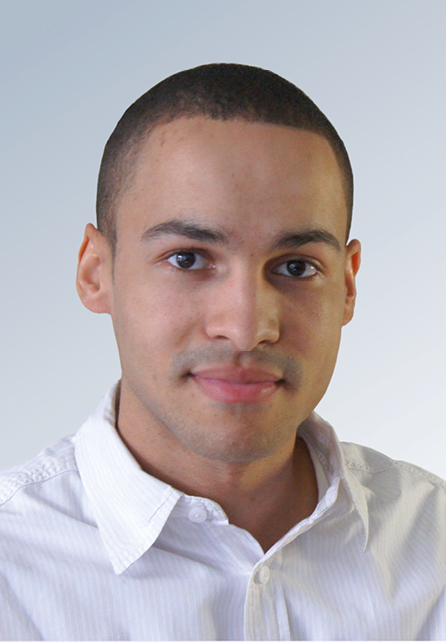}}]
    {Benedict Flade} studied simulation and control of mechatronic systems and received a master’s degree from TU Darmstadt, Germany. 
    Since 2016, he is working as a scientist at the Honda Research Institute Europe GmbH. 
    His research aims to improve both terrestrial and aerial intelligent transportation systems. 
    Specifically, his research interests cover environment representation concepts, digital cartography, vehicle localization systems, and absolute and map-relative positioning approaches.
    \end{IEEEbiography}%
    \begin{IEEEbiography}[{\includegraphics[width=0.9in,keepaspectratio]{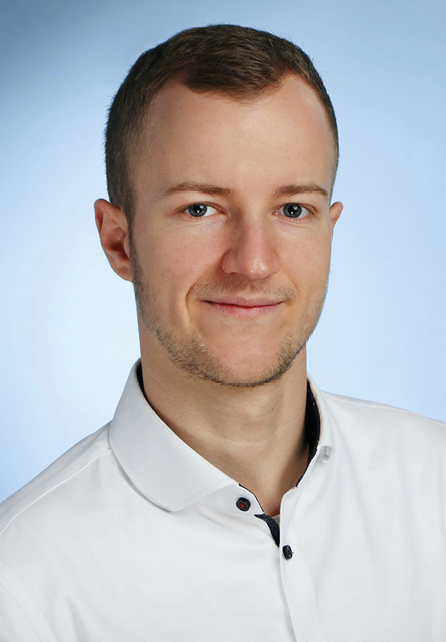}}]
    {Daniel Ochs} graduated from the Technical University of Darmstadt, Germany, with a bachelor’s degree in Business Informatics and a master’s degree in Computer Science. 
    Since 2022, he is pursuing his Ph.D. with the Artificial Intelligence and Machine Learning Lab. 
    His fields of interest are Neuro-Symbolic AI and multimodal models for Visual Question Answering. 
    \end{IEEEbiography}%
    \begin{IEEEbiography}[{\includegraphics[width=0.9in,keepaspectratio]{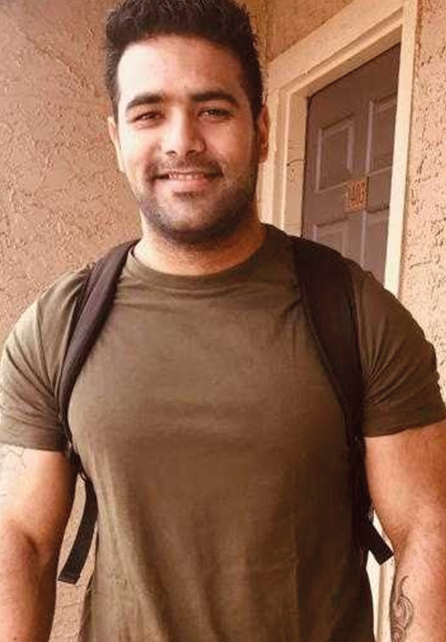}}]
    {Devendra Singh Dhami} joined the Uncertainty in Artificial Intelligence group at TU Eindhoven as an Assistant Professor in 2023.
    Before moving to Eindhoven, he completed his doctorate at the University of Texas at Dallas in 2020.
    Afterward, he spent three years as a postdoctoral researcher at TU Darmstadt in Germany and became a junior research group leader at the Hessian Center for Artificial Intelligence in 2022.
    Devendra's research interests currently focus on successfully incorporating causality and reasoning into deep learning systems.
    \end{IEEEbiography}%
    \begin{IEEEbiography}[{\includegraphics[width=0.9in,keepaspectratio]{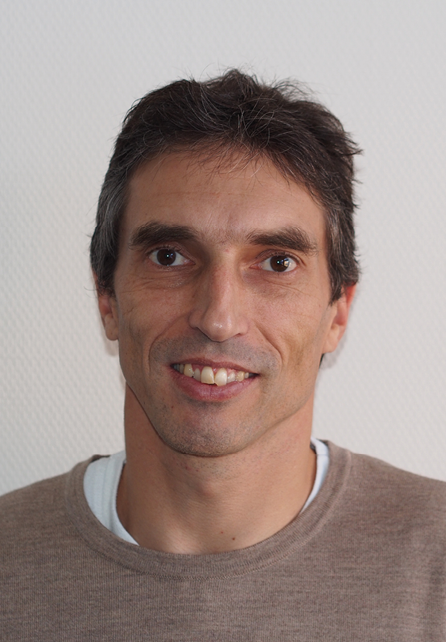}}] 
    {Julian Eggert} received his Ph.D. degree in physics from the Technical University of Munich. 
    In 1999, he joined Honda R\&D (Germany) and, in 2003, the Honda Research Institute, where he is currently a Chief Scientist and leads projects in artificial cognitive systems with applications in car and robotics domains. 
    His fields of interest are generative models for perception, large-scale models for visual processing and scene analysis, and semantic environment models for context-embedded reasoning, situation classification, risk prediction, and behavior planning. 
    \end{IEEEbiography}%
    \begin{IEEEbiography}[{\includegraphics[width=0.9in,keepaspectratio]{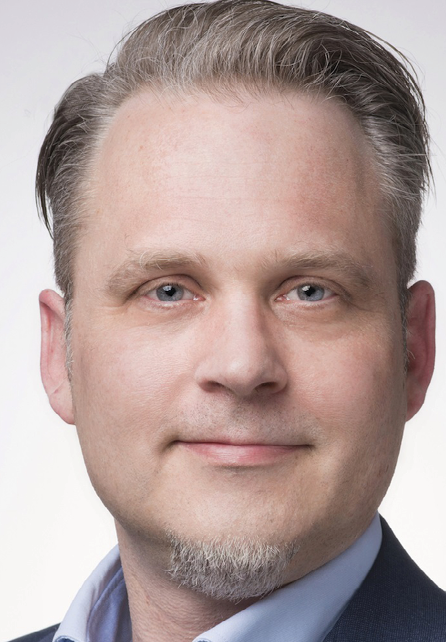}}]
    {Kristian Kersting} is a Full Professor at the Computer Science Department of the TU Darmstadt University, Germany. 
    He is the head of the Artificial Intelligence and Machine Learning lab, a member of the Centre for Cognitive Science, a faculty of the ELLIS Unit Darmstadt, and the founding co-director of the Hessian Center for Artificial Intelligence. 
    After receiving his Ph.D. from the University of Freiburg in 2006, he was with the MIT, Fraunhofer IAIS, the University of Bonn, and the TU Dortmund. 
    His main research interests are statistical relational artificial intelligence and deep (probabilistic) programming and learning.
    \end{IEEEbiography}%
\end{document}

%% file: header.tex
\usepackage{cite}
\usepackage{amsmath,amssymb,amsfonts}
\usepackage{algorithmic}
\usepackage{graphicx}
\usepackage{textcomp}
\usepackage{balance}
\usepackage{epstopdf}
\usepackage[normalem]{ulem}
\usepackage{upgreek}
\usepackage{comment}
\usepackage{hyperref}
\usepackage{tikz}
\usepackage{lipsum}
\usepackage[formats]{listings}
\usepackage{subcaption}
\usepackage{multirow}
\usepackage{siunitx}
\usepackage[frozencache,cachedir=.]{minted}
\usepackage[color=black,opacity=1,angle=0,scale=1]{background}

%% file: content/0_abstract.tex
Advanced Air Mobility (AAM) is a growing field that demands accurate and trustworthy models of legal concepts and restrictions for navigating Unmanned Aircraft Systems (UAS).
In addition, any implementation of AAM needs to face the challenges posed by inherently dynamic and uncertain human-inhabited spaces robustly.
Nevertheless, the employment of UAS beyond visual line of sight (BVLOS) is an endearing task that promises to significantly enhance today's logistics and emergency response capabilities.
Hence, we propose Probabilistic Mission Design (ProMis), a novel neuro-symbolic approach to navigating UAS within legal frameworks.
ProMis is an interpretable and adaptable system architecture that links uncertain geospatial data and noisy perception with declarative, Hybrid Probabilistic Logic Programs (HPLP) to reason over the agent's state space and its legality.
To inform planning with legal restrictions and uncertainty in mind, ProMis yields Probabilistic Mission Landscapes (PML).
These scalar fields quantify the belief that the HPLP is satisfied across the agent's state space.
Extending prior work on ProMis' reasoning capabilities and computational characteristics, we show its integration with potent machine learning models such as Large Language Models (LLM) and Transformer-based vision models.
Hence, our experiments underpin the application of ProMis with multi-modal input data and how our method applies to many AAM scenarios.


%% file: content/1_introduction.tex
\section{Introduction}
\label{sec:introduction}

\IEEEPARstart{M}{ission} 
design for intelligent transportation systems is tied to several challenges, mandating strict adherence to safety criteria and public laws when traversing human-inhabited spaces. 
Hence, robust representations of the mandated constraints for navigating uncertain environments are highly sought after.
Not only must they be interpretable, but also allow adaptations to regulatory changes and reasoning on an agent's actions' compliance.
These demands are all the more pressing when employing \textbf{Unmanned Aircraft Systems~(UAS)}, as they pose a safety hazard to uninvolved bystanders.

Although policies for \textbf{Advanced Air Mobility~(AAM)} scenarios have emerged, frameworks that capture these regulations in planning and control while considering the uncertainty of background knowledge and perception, e.g., maps and deep learning models, have remained absent.

\input{figures/pml}

\IEEEpubidadjcol

To bridge this technical gap, we believe that neuro-symbolic systems, integrating declarative, white-box models with robust machine learning systems, are a promising foundation for future AAM systems.
Hence, we propose \textbf{Probabilistic Mission Design (ProMis)}~\cite{Kohaut2023} as a foundation for neuro-symbolic UAS.
In contrast to research focusing on informing the agent's decisions purely based on energy, time, or collision avoidance criteria~\cite{chenMultiDrone, guanEfficientUAV, hohmann2023three}, ProMis generates \textbf{Probabilistic Mission Landscapes~(PML)} as shown in Figure~\ref{fig:pml} to quantify the probability of satisfying local AAM regulations as a trustworthy navigation platform for downstream tasks.

\input{listings/problog_example}

ProMis combines Hybrid Probabilistic Logic Programs~\cite{nittihybrid} (as illustrated in Listing~\ref{listing:spatial_relations}) with semantically annotated geographic data (as exemplified in Listing~\ref{listing:overpass_ql}) and neural perception.
We show how ProMis provides a neuro-symbolic framework that unifies multi-modal data sources, mission settings, and constraints into an interpretable and adaptable basis for safely employing UAS beyond visual line of sight.
More specifically, we make the following key contributions to intelligent transportation systems:
\begin{itemize}
    \item We show how uncertain environment information from crowd-sourced map data and neural perception can be compiled into neuro-symbolic spatial relations as a vocabulary for expressing AAM rules in first-order logic, resulting in ProMis' ability to generate Probabilistic Mission Landscapes.
    \item We demonstrate how Transformer-based vision models such as the ChangeFormer architecture~\cite{ChangeFormer} can be integrated with ProMis, expanding on ProMis' expressivity and showcasing its potential for future neural-perception systems to be incorporated into mission design.
    \item We show how a general-purpose \textbf{Large Language Model~(LLM)} can be employed to translate the natural language prompts of an operator into ProMis code, illustrating how a neural system can bridge the gap between non-experts and our method.
\end{itemize}
By these extensions and expanded evaluation compared to prior work, we pave the way for future intelligent transportation systems to be guided in their actions by declarative, probabilistic programs to respect a set of interpretable rules.
Furthermore, we provide an open-source implementation of the ProMis framework at~\url{https://github.com/HRI-EU/ProMis}.

%% file: figures/pml.tex
\begin{figure}
    \centering
    \includegraphics[width=0.85\linewidth]{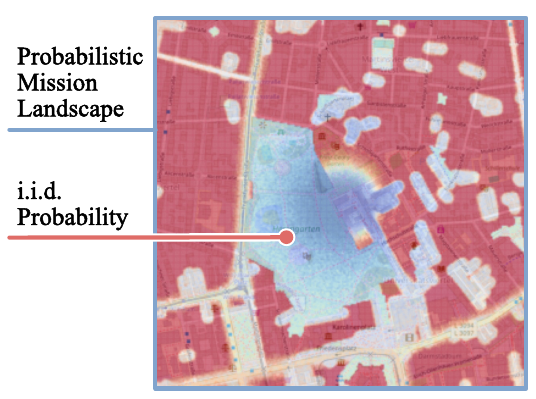}
    \caption{
        \textbf{Probabilistic Mission Landscapes:}
        Through ProMis, each point in the agent's state space is associated with an independent and identically distributed (i.i.d.) probability of satisfying the modeled navigation constraints under uncertainty (from high probability in blue to low in red).
        Due to the independence of each point, samples can be computed in parallel and interpolated into a scalar field. 
    }
    \label{fig:pml}
\end{figure}

%% file: listings/problog_example.tex
\begin{listing}[t]
    \centering
    \begin{minted}
    [
        frame=none,
        autogobble,
        fontsize=\footnotesize
    ]{prolog}
        % Spatial relations from statistical evaluation
        distance(x0, building) ~ normal(20, 0.5).
        0.9::over(x0, primary).
        ...

        % Spatial relations from neural sensors 
        0.15::change(x0).
        ...

        % Satisfying circumstances for the mission
        landscape(X) :- 
            \+ change(X), distance(X, building) > 50;
            over(X, park), distance(X, primary) < 15.
    \end{minted}
    \caption{
        Modeling probabilistic flight requirements in ProMis as Hybrid Probabilistic Logic Program.
        The parameters may be estimated via statistical evaluation of uncertain map data or through a deep learning model on, e.g., image data.
    }
    \label{listing:spatial_relations}
\end{listing}

%% file: content/2_related_work.tex
\section{Related Work}
\label{sec:related_work}

\subsection{Public Regulations in Unmanned Aerial Mobility}

In Europe, the advancement of UAS and \textbf{Urban Air Mobility~(UAM)} is significantly influenced by the Single European Sky Air Traffic Management Research initiative. 
When they introduced their first Master Plan in 2009, the concept of AAM was not considered, with drones merely mentioned in passing.
However, with subsequent editions of the Master Plan, particularly the 2015 release~\cite{undertaking2016european} and the 2017 Drone Outlook Study~\cite{sesar_2017}, the integration of UAS into the European airspace gained momentum.

Besides substantial funding allocated to drive development, operational restrictions have been introduced concurrently to ensure the safety of UAS operations across urban, industrial, and rural environments.
Notably, the EU regulation 2019/947~\cite{Ec2019}, implemented in 2019, underscores the coexistence of unmanned and manned aircraft within shared airspace, emphasizing rigorous risk assessments, such as through the Specified Operations Risk Assessment~(SORA method)~\cite{Easa2019}.

ProMis is motivated by these developments, aiming to allow the UAS to know about the local rules and to reason about their satisfaction.
While prior work on ProMis has merely considered uncertainty in geographic data, e.g., crowd-sourced maps, this work moves one step further to incorporate multi-modal data through neural components into ProMis' quantification of compliance with public regulations. 

\input{listings/overpass_ql}

\subsection{Automated Risk Assessment}
Recent research endeavors have enhanced safety analysis and risk assessment methodologies.
For instance, Rothwell and Patzek~\cite{RothwellPatzek2019} have contributed by employing satisfiability checks on symbolic models to verify and improve mission planning for UAS.
In contrast, Rakotonarivo et al.~\cite{Rakotonarivo2022} highlight the importance of fitting the safety analysis and risk assessment output to map or environmental data, proposing interactive visual representations and data exploration. 

\input{figures/architecture}

Similarly, Primatesta et al.~\cite{Primatesta2020} and Raballand et al.~\cite{Raballand2021} have introduced approaches that integrate visual maps with risk models to calculate and visually represent risks associated with UAS operations. 
These models prioritize risk assessment based on formal frameworks, identifying potential hazards and safety concerns, especially ground casualties and transportation network disruptions.

Even if humans are in complete control, unawareness of situations may quickly lead to life-threatening scenarios, with controlled flight into terrain accounting for a significant proportion of aviation fatalities~\cite{cfit_aviation}.
For this reason, rich visualizations of the mission in progress have been developed to avoid blind spots in decision-making~\cite{visualization_aviation}.
By providing insight into the mission and its environment from information sources, operators can react to threats and optimize their behavior~\cite{gis_aviation, weather_information_impact}.

With ProMis, we improve safety in AAM by presenting a novel neuro-symbolic approach to automated risk assessment in UAS, yielding visual representations and planning objectives representing risks during flight analogous to prior work.
In contrast to the state-of-the-art, we demonstrate how neural components can be integrated with ProMis to create or extend these encodings from natural language prompts and how deep learning models incorporate multi-modal input data into the reasoning process. 

\subsection{Neuro-Symbolic Systems}

Symbolic reasoning systems are a suitable basis for an adaptable and interpretable system to formalize and verify constrained behavior.
One of the earliest programmatic reasoning systems in \textbf{First-Order Logic~(FOL)} is Prolog~\cite{colmerauer1990introduction}, developed in 1972 by Alain Colmerauer.
Prolog has inspired numerous applications, from natural language processing~\cite{nittihybrid} to robotics~\cite{robot_prolog}.
Extensions that embrace uncertainties in formal logic, like Bayesian Logic Programs~\cite{bayesian_logic} and Probabilistic Logic Programs~\cite{problog,inference_in_plp}, have been introduced to allow for weighted inference in FOL models.

While they are not formulated for end-to-end learning in tandem with artificial neural networks, recent models like DeepProblog~\cite{deepproblog} and SLASH~\cite{slash} have been introduced, tightly integrating deep learning models with symbolic reasoning.
Such neuro-symbolic systems aim to combine the strengths of deep learning architectures for tasks such as noisy perception with symbolic, formally well-justified, and interpretable systems in an end-to-end learnable fashion.

In this work, by employing \textbf{Hybrid Probabilistic Logic Programs~(HPLP)}~\cite{nittihybrid}, accounting for discrete and continuous distributions, in tandem with neural components such as LLMs and Transformer-based vision models, we close the gap between state-of-the-art neuro-symbolic systems and UAS navigation.
Hence, with ProMis, we present a novel approach for neuro-symbolic UAS operations, paving the way towards end-to-end learned, adaptable, and interpretable mission design.

%% file: listings/overpass_ql.tex
\begin{listing}[t]
    \centering
    \begin{minted}
    [
        frame=none,
        autogobble,
        fontsize=\footnotesize
    ]{c}
    // Output format and timeout for request
    [out:json][timeout:25];

    // Requested ways and relations within a bounding box
    (
        way["highway"="primary"]({{bbox}});
        way["highway"="secondary"]({{bbox}});
        relation["natural"="bay"]({{bbox}});
        relation["building"]({{bbox}});
    );

    // Return retrieved data
    out body; >; out skel qt;
    \end{minted}
    \caption{
        The Overpass Query Language for querying semantically annotated geographic data from OpenStreetMap (OSM), a crowd-sourced database with an open license and widespread adoption.
    }
    \label{listing:overpass_ql}
\end{listing}

%% file: figures/architecture.tex
\begin{figure*}
    \centering
    \includegraphics[width=\textwidth]{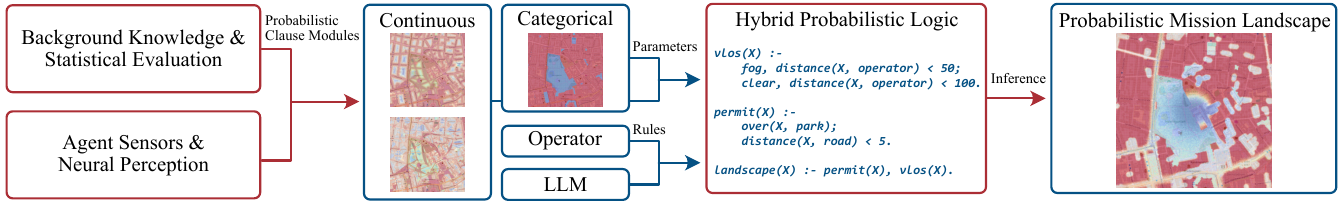}
    \caption{
        \textbf{The Probabilistic Mission (ProMis) system architecture:}
        Probabilistic Clause Modules map the agent' provided environment data from statistical (see Section~\ref{sec:nesy_relations}) and neural (see Section~\ref{sec:neural_relations}) models into continuous and categorical distributions over the agent's navigation space.
        Combined with the mission rules encoded by an operator or via an LLM, a Hybrid Probabilistic Logic Program is built to infer the satisfying state space as a Probabilistic Mission Landscape.
    }
    \label{fig:architecture}
\end{figure*}

%% file: content/3_methods.tex
\section{Methods}
\label{sec:methods}

We now present (i) the architecture of ProMis, (ii) the generation of neuro-symbolic spatial relations from semantically annotated map data and neural perception, integrating multi-modal input data of AAM settings with Hybrid Probabilistic Logic Programs (HPLP), and (iii) the inference of Probabilistic Mission Landscapes (PML) as scalar fields of probabilities across the agent's state space.

\subsection{An Architecture for Probabilistic Mission Design}

With ProMis, we aim at a \underline{Pro}babilistic \underline{Mis}sion design flow, integrating the multimodal data and needs of the involved AAM stakeholders.
An illustration of ProMis' architecture is shown in Figure~\ref{fig:architecture}.
We consider three stakeholders contributing data to this pipeline.

First, the UAS manufacturer creates an agent that is equipped with a set of \textit{sensors} and \textit{neural perception} modules, as well as access to \textit{background knowledge}, e.g., a map, with accuracy information for \textit{statistical evaluations}.
We incorporate this data through \textbf{Probabilistic Clause Modules~(PCM)} into a set of neuro-symbolic spatial relations (see Section~\ref{sec:neural_relations}).

Second, a public body regulates UAS and flight categories.
This includes the classifications and categorizations of autonomous agents and their operations, giving rise to a need to develop safe and legally compliant mission models. 
In addition to creating \textit{rules} for public spaces, they provide data on critical infrastructure or areas that privately operated agents should not enter, e.g., geofences protecting the space around airports as part of the \textit{background knowledge}.

Third, the UAS operator determines the mission's target and additional \textit{rules} reflecting their requirements on top of official regulations.
As operators may not possess the necessary engineering expertise to encode their requirements manually, neural language processing provided by an LLM may be employed (see Section~\ref{sec:llm}).

Based on these sets of \textit{parameters} and \textit{rules}, \textit{inference} can be run on an HPLP for generating a PML over the agent's state space (see Section~\ref{sec:pml}).
The PML may then act as an objective for autonomous planning by the UAS and for informing the operator about the safety and compliance of their operations.

\subsection{Neuro-Symbolic Spatial Relations}
\label{sec:nesy_relations}

To enable probabilistic reasoning about the compliance of an agent's state with a given set of regulations, a vocabulary of spatial relations is necessary to express the relevant legal concepts.
For example, in the open category for low-risk UAS operations under the European Union Aviation Safety Agency~\footnote{\url{https://www.easa.europa.eu/en/domains/drones-air-mobility/operating-drone/open-category-low-risk-civil-drones}}, a UAS must never fly \textit{over} an assembly of people or keep a minimum \textit{distance} from urban areas.
While the former is a categorical expression (yes or no), the latter references a continuous domain (distance in meters).
Hence, in ProMis, we formalize and compute the parameters for hybrid (categorical and continuous) probabilistic spatial relations through neural and symbolic models as a basis for expressing AAM and operator rules (as seen in Listing~\ref{listing:spatial_relations}).

First, let us consider the statistical evaluation of semantically annotated, uncertain map data.
Let a map $\mathcal{M} = (\mathcal{V}, \mathcal{E}, \tau)$ be the triple of vertices $\mathcal{V}$, edges $\mathcal{E}$ and annotator function $\tau$.
Each vertex $\mathbf{v} \in \mathcal{V}$ is located in a $D$-dimensional space $\mathbb{R}^D$ of Cartesian coordinates, and each edge $\mathbf{e} \in \mathcal{E}$ is a pair of connected vertices.
If two vertices are connected via the edges in the map, we consider them part of the same feature. 
Furthermore, the annotator function $\tau(\mathbf{v}) = \mathcal{T}_\mathbf{v}$ produces the set $\mathcal{T}_\mathbf{v}$ of semantics assigned to $\mathbf{v}$.
For example, if $\mathbf{v}$ is a node within a polyline defining the main street of a city, one may set $\tau(\mathbf{v}) = \{\text{primary}\}$.

To address possible inaccuracies of the map data, e.g., due to the use of low-cost sensors or crowd-sourced data, we employ a stochastic error model analogously to prior work~\cite{flade2021error}.
Hence, we consider for each $\mathbf{v}_{i, j} \in \mathcal{V}$, being the $j$-th vertex of the $i$-th feature, the following stochastic affine map generating $N \in \mathbb{N}$ samples:
\begin{align*}
    \mathbf{\Phi}^{(n)} &\sim \phi_i \tag*{(Transformation)} \\
    \mathbf{t}^{(n)} &\sim \kappa_i \tag*{(Translation)} \\
    \mathbf{v}^{(n)}_{i,j} &= \mathbf{\Phi}^{(n)} \cdot \mathbf{v}_{i, j} + \mathbf{t}^{(n)} \tag*{(Generation)}
\end{align*}
Here, $\phi_i$ and $\kappa_i$ are feature-wise distributions of linear maps $\mathbf{\Phi}^{(n)}$, e.g., rotation, scaling or shearing, and translations $\mathbf{t}^{(n)}$.

Consider the previous example of \textit{over} and \textit{distance} as spatial relations in encoding AAM regulations. 
With $\mathcal{M}^{(n)}$ being generated by taking the $n$-th sample of each vertex, this collection of randomized maps will now allow us to fit the parameters of such relations empirically.

Let $j(\mathcal{M}^{(n)}, \mathbf{x}, t)$ be a deterministic function evaluating a spatial relation on map $\mathcal{M}^{(n)}$ at location $\mathbf{x} \in \mathbb{R}^D$ limited to features for which $\tau(\mathbf{v}) = t$.
Through this sampling process, we can empirically estimate statistical moments, e.g., mean and variance, with respect to the chosen relation $j$, location $\mathbf{x}$, and semantics $t$:%
\begin{align}
    \widehat{\mu_j} &= \frac{1}{N} \sum_n j(\mathcal{M}^{(n)}, \mathbf{x}, t) \\
    \widehat{\sigma^2_j} &= \frac{1}{N - 1} \sum_n (j(\mathcal{M}^{(n)}, \mathbf{x}, t) - \widehat{\mu_j})^2
\end{align}%
For example, assume \textit{distance}$(\mathbf{x}, t) \sim \mathcal{N}(\mu_d, \sigma^2_d)$ to be a normally distributed random variable with the deterministic function $d$.
That is, the function $d(\mathcal{M}^{(n)}, \mathbf{x}, t)$ shall compute the Euclidean distance from $\mathbf{x}$ to the closest feature in $\mathcal{M}^{(n)}$.
An illustration of this process in the case of the \textit{distance} relation for a single \textit{road} segment with different error models is shown in Figure~\ref{fig:map_variations}.

While one can immediately set the parameters of the normal distribution as $\mu_d = \widehat{\mu_d}$ and $\sigma_d^2 = \widehat{\sigma_d^2}$, employing, e.g., the method of moments or maximum likelihood is possible for other distributions.

Of course, not every spatial relation can be extracted from map data.
Instead, if, for example, images need to be analyzed, it is necessary to employ neural components instead.
This means that rather than a sampling-based evaluation of background knowledge, a deep learning model is trained to estimate the distribution of the spatial relation directly.
We explore this further in our experiments in Section~\ref{sec:neural_relations}.

\subsection{Hybrid Probabilistic Logic Programs for Aerial Mobility}
\label{sec:hplp_for_am}

HPLPs intertwine formal logic with probability theory over hybrid relational spaces, i.e., categorical and continuous distributions~\cite{nittihybrid}.
While HPLPs are domain-agnostic models in which probabilistic logic can answer a wide range of queries, we focus on treating data for UAV navigation.
For instance, they may encode facts and relations regarding background knowledge about an operator's license, a UAS's categorization, local flight restrictions, and manufacturer specifications.

\input{figures/map_variations}

An HPLP consists of at least one \textit{clause}, with each clause being made up of a \textit{head}, \textit{body}, and \textit{distribution}.
Consider the following two clauses.
\begin{align}
    \label{eq:clauses}
    p \ :: \ &r_1(a_1, \ldots, a_n) \ \text{:-} \ l_1,\ \ldots,\ l_m. \tag*{(Categorical)} \\
    &r_2(a_1, \dots, a_i) \sim p(\mathbf{\theta}) \ \text{:-} \ l_1,\ \ldots,\ l_j. \tag*{(Continuous)}
    \label{eq:distributional_clauses}
\end{align}
In the first case, the head $r_1$ is true with a probability $p$ given that all the \textit{literals} $l_k$ in the body are true.
Analogously, in the second case, head $r_2$ is distributed according to density $p(\mathbf{\theta})$ with parameters $\mathbf{\theta}$ if its body is true.
If the right-hand side is empty, the head is regarded as a fact and distributed independently of any other symbols.

This model of knowledge in uncertain domains is especially interesting when considering ProMis' spatial relations as formalized in Section~\ref{sec:nesy_relations}.
For example, let us formalize our running examples of \textit{distance} and \textit{over} as HPLP clauses:
\begin{align}
    \label{eq:example_clause_over}
    &0.25 \ :: \ \text{over}(\text{x0}, \text{assembly\_of\_people}). \\
    &\text{distance}(\text{x0}, \text{urban\_area}) \sim normal(100, 1).
    \label{eq:example_clause_distance}
\end{align}
This means that at the location referenced with the term x0, there is an assembly of people with a probability of $0.25$, and the distance to the nearest urban area is expected to be about $\SI{100}{\meter}$.

In contrast to pure logic programming languages, where a query will either be evaluated as true or false, probabilistic logic programs can be employed as generative probabilistic models.
Combining the distributions obtained from Section~\ref{sec:nesy_relations} with clauses  provided by a public body or operator, one can query safety-critical questions about the states of the UAS navigation space.

To do so, ProMis generates code in the form of Equations~\ref{eq:example_clause_over} and \ref{eq:example_clause_distance} to be employed as probabilistic vocabulary for stating AAM policies and mission restrictions, providing the operator and agent with a tool to quantify the probability with which safety-critical requirements are satisfied.

\input{figures/distributional_atoms}

\label{sec:uncertain_maps}

\subsection{Probabilistic Mission Landscapes}
\label{sec:pml}

ProMis' main contribution is quantifying the probability that agent states satisfy local flight regulations.
To do so, we propose \textbf{Probabilistic Mission Landscapes~(PML)} $P(c | \mathbf{x})$ as scalar fields of probability.
Here, we refer to $P(c | \mathbf{x})$ as the probability that the agent's set of probabilistic rules, as derived in Sections~\ref{sec:nesy_relations} and~\ref{sec:hplp_for_am}, in a sense, its \textit{constitution} $c$, is satisfied for state $\mathbf{x}$.

Given the appropriate background knowledge, sensors, and neural components as described before, ProMis generates the agent's PML in three steps.
First, the neuro-symbolic spatial relations are evaluated by statistical evaluation of the agent's background knowledge and neural evaluation of sensor data (Section~\ref{sec:neural_relations}).
Second, through ProMis' PCMs, the obtained parameters are written into an HPLP, which is combined with operator or LLM-provided clauses specifying the mission restrictions (Section~\ref{sec:hplp_for_am}).
Finally, the HPLP is queried for the \textit{landscape(X)} clause, which is required to be defined to summarize all requirements into a single symbol (see Listings~\ref{listing:spatial_relations} and~\ref{listing:uam_model} for comparison).

While laying out the complete inference procedure for HPLPs is out of scope for this work (we refer the interested reader to Hybrid Relational ProbLog~\cite{nittihybrid}), we outline the core idea.
An HPLP as a first-order logic program needs first to be \textit{grounded}, i.e., a program in which only ground atoms (variable-free, not further decomposable symbols) $\mathcal{A}$ remain.
Then, the grounded program is \textit{solved}, meaning all models $\mathcal{U}$ for which the queried atom is true are enumerated.
The probability of a query is then the sum of probabilities of the individual models, which in turn are the product of the probabilities of the ground atoms being assigned the respective Boolean value:
\begin{align}
    P(c | \mathbf{x}) = \sum_{u \in \mathcal{U}} \prod_{a \in \mathcal{A}} P(a = u(a))
\end{align}
Note that the probability of the continuous distributions is either achieved via the Cumulative Distribution Function or through sampling, depending on whether a closed-form solution is available.

\input{figures/changeformer}

%% file: figures/map_variations.tex
\begin{figure}[t]
    \begin{subfigure}[t]{0.16\textwidth}
        \includegraphics[width=\textwidth]{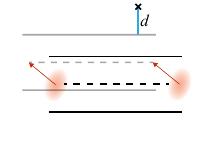}
        \caption{Translation}
    \end{subfigure}\hfill%
    \begin{subfigure}[t]{0.16\textwidth}
        \includegraphics[width=\textwidth]{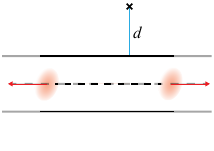}
        \caption{Scaling}
    \end{subfigure}\hfill%
    \begin{subfigure}[t]{0.16\textwidth}
        \includegraphics[width=\textwidth]{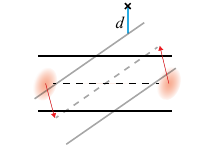}
        \caption{Rotation}
    \end{subfigure}
    \caption{
        \textbf{Sampling spatial relations from uncertain maps:}
        With an expectation of a feature's true location (black road segment) and the respective error parameters, one can generate map variations (gray road segments) to estimate the parameters, e.g., of the distance $d$ from a reference point $\mathbf{x}$.
    }
    \label{fig:map_variations}
\end{figure}

%% file: figures/distributional_atoms.tex
\begin{figure*}
    \centering
    \begin{subfigure}{0.27\textwidth}
        \includegraphics[width=\textwidth]{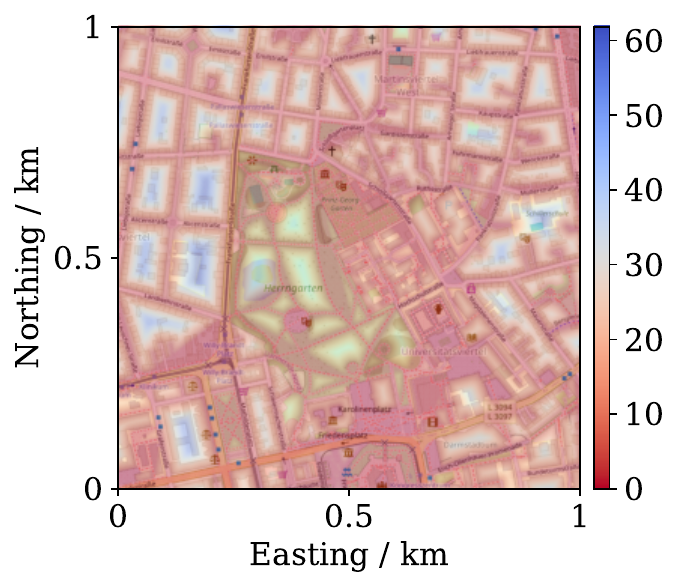}
        \caption{$\mu_d$ of \textit{distance(X, road)}}
    \end{subfigure}
    \hfill
    \begin{subfigure}{0.23\textwidth}
        \includegraphics[width=\textwidth]{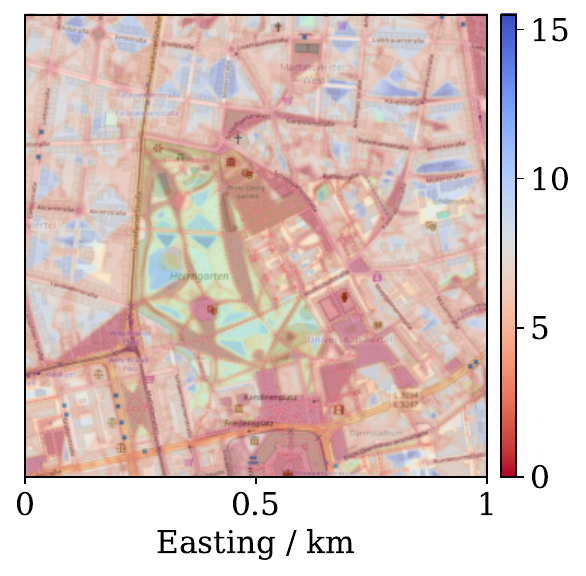}
        \caption{$\sigma^2_d$ of \textit{distance(X, road)}}
    \end{subfigure}
    \hfill
    \begin{subfigure}{0.235\textwidth}
        \includegraphics[width=\textwidth]{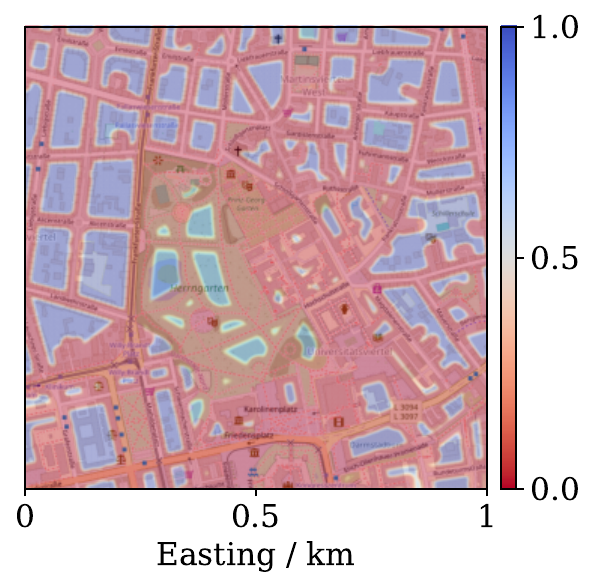}
        \caption{$P(distance(X, road) > 15)$}
    \end{subfigure}
    \hfill
    \begin{subfigure}{0.235\textwidth}
        \includegraphics[width=\textwidth]{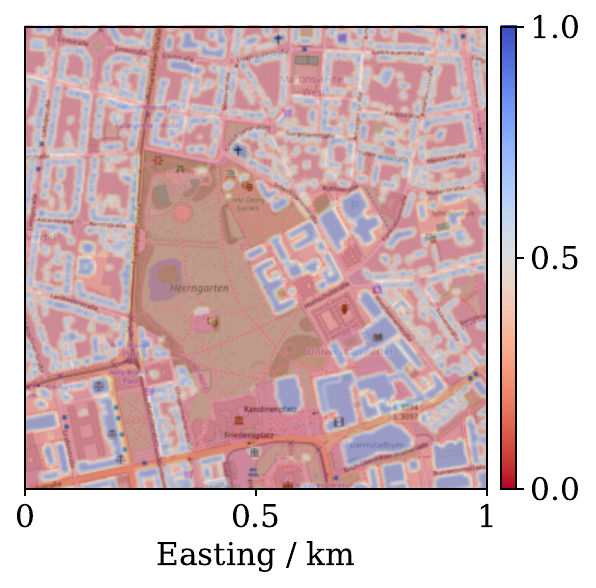}
        \caption{$P(over(X, building))$}
    \end{subfigure}
    \caption{
        \textbf{Hybrid probabilistic spatial relations from statistical evaluation:}
        We show the mean (a) and variance (b) of \textit{distance(X, road)}.
        Through integration, one obtains the probability of a regulatory constraint being satisfied, e.g., keeping a distance of over $\SI{15}{\meter}$ to roads (c).
        Finally, (d) shows \textit{P(over(X, building))}, which as a categorical distribution can be directly taken from the estimated parameters.
        By associating the parameters with points in state-space, they are translated into HPLP clauses as in Listing~\ref{listing:spatial_relations}.
        Colors have been made transparent to reveal the map data employed as background knowledge.
    }
    \label{fig:distributional_atoms}
\end{figure*}

%% file: figures/changeformer.tex
\begin{figure*}
    \centering
    \begin{subfigure}{0.245\textwidth}
        \includegraphics[width=\textwidth]{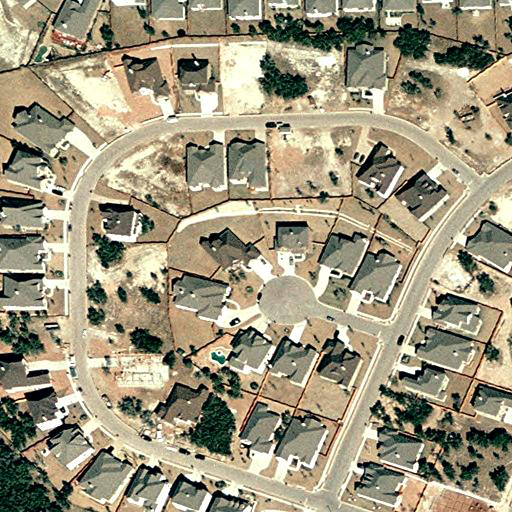}
        \caption{Before}
    \end{subfigure} 
    \begin{subfigure}{0.245\textwidth}
        \includegraphics[width=\textwidth]{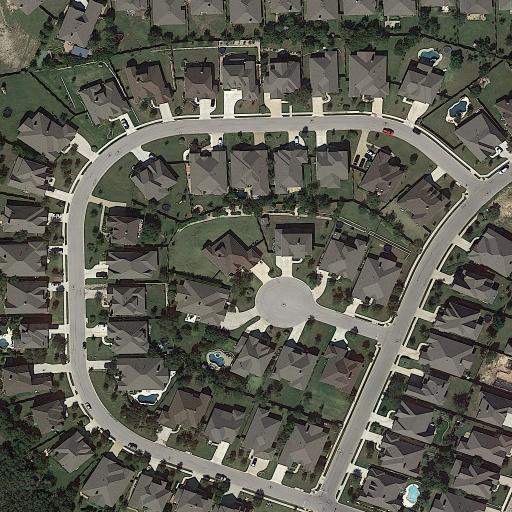}
        \caption{After}
    \end{subfigure}
    \vspace{0.35em}
    \begin{subfigure}{0.245\textwidth}
        \includegraphics[width=\textwidth]{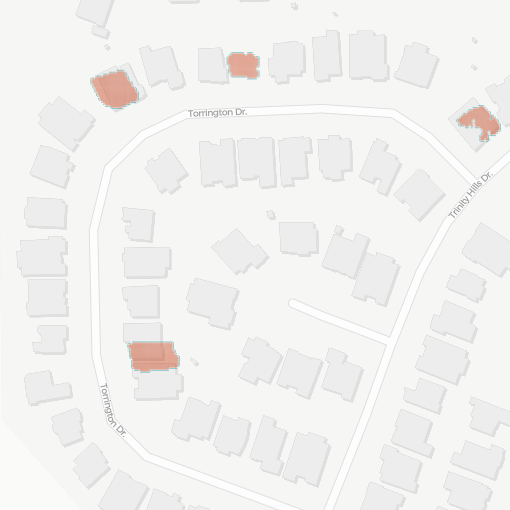}
        \caption{$P(change(X))$}
    \end{subfigure} 
    \begin{subfigure}{0.245\textwidth}
        \includegraphics[width=\textwidth]{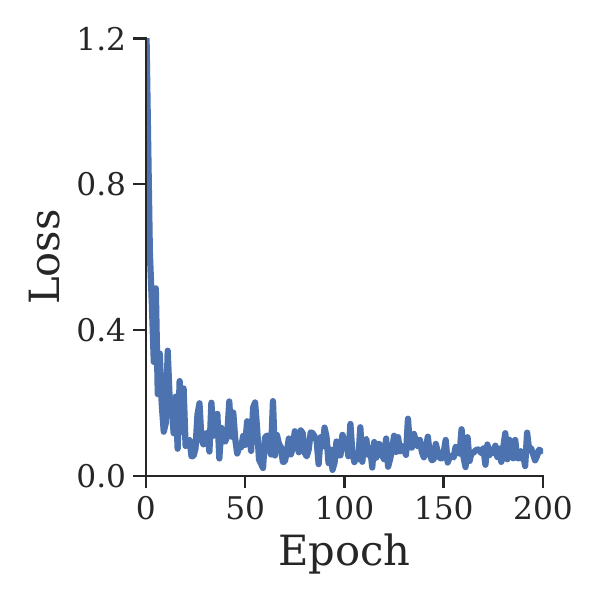}
        \caption{Training}
    \end{subfigure}
    \caption{
        \textbf{Hybrid probabilistic spatial relation from neural sensing:} 
        Before (a) and after (b), images taken by a satellite provide the input for the ChangeFormer, resulting in the parameters of \textit{change(X)} (c) after training the model on ground-truth data (d).
        By associating the prediction with points in state-space, it is translated into HPLP clauses as in Listing~\ref{listing:spatial_relations}.
        Note how the ChangeFormer is, on the one hand, not perfect and missed some of the construction sites in the west, but, on the other hand, identified a building that is not part of OpenStreetMap at the time of writing (center-top).
    }
    \label{fig:changeformer}
\end{figure*}

%% file: content/4_experiments.tex
\section{Experiments}
\label{sec:results}

\subsection{Spatial Relations from Crowd-Sourced Map Data}
\label{sec:distance_over_exp}

We begin our experiments by demonstrating the computation of two spatial relations from crowd-sourced OpenStreetMap data.
Namely, we consider the \textit{over(x, building)} and \textit{distance(x, road)} spatial relations from our running examples on the statistical evaluation of inaccurate map data.

To this end, we query data from OpenStreetMap using Overpass, akin to Listing~\ref{listing:overpass_ql}.
For the map generation process, we assume the geometry to have a simple translational error, i.e., we set the transformation $\phi_i$ to be the identity matrix and the translation $\kappa_i = \mathcal{N}(0, \text{diag}(10, 10))$ for all features.

For this experiment, we place the origin of the mission space in Darmstadt, Germany (latitude~$=49.878091$, longitude~$=8.654052$).
Furthermore, we draw the sampling locations on a $1000 \times 1000$ regular grid and generate $N = 50$ maps in a $\SI{1}{\kilo\meter\squared}$ area.
The resulting spatial relations are visualized in Figure~\ref{fig:distributional_atoms}.

\subsection{Neural Perception in Probabilistic Mission Design}
\label{sec:neural_relations}

Beyond statistical analysis of, e.g., geographic information systems, neural perception is an essential ingredient in ProMis' ability to reason based on the agent's immediate surroundings. 
Hence, in this section, we consider the classification output of a Transformer-based vision model for inclusion in the generated HPLP.
While the potential for neural perception is vast, ranging from locating bystanders on the ground to detecting forest fires, we will consider change detection from satellite images as an example of translating neural perception into spatial relations. 
For this, we have trained a ChangeFormer~\cite{ChangeFormer} following the training approach described in~\cite{rs16020266}, employing the LEVIR-CD dataset~\cite{levir} to learn to detect salient changes from satellite images. 

The LEVIR-CD dataset is an urbanization monitoring dataset featuring before and after image pairs of developing suburbs, e.g., allowing the ChangeFormer to detect construction sites and newly finished buildings. 
Further, the ChangeFormer is a Siamese-like architecture that applies feature extraction on four scales to both the before- and after-images. 
Next, the features are compared via a learned difference module, and the resulting comparison features are fed through fully connected up-sampling layers to produce the change map. 
Hence, each pixel of the change map corresponds to a categorical distribution indicating the probability of change.

We chose this example for two reasons:
First, regardless of semantics, generating HPLP relations from the ChangeFormer output is analogous to any other neural perception that can be mapped to the agent's navigation space.
Second, besides encoding the ChangeFormer's output into the HPLP and providing a more expressive language for mission design, its semantics can be leveraged to detect outdated map data.
Hence, as seen in Figure~\ref{fig:changeformer}, neural perception is not only a more dynamic form of obtaining mission-related knowledge but also opens up the ability to reason on the differences between multiple modes of input data.
Of course, the same reasoning goes in reverse, with the map data being more complete than the ChangeFormer's detection.

Listing~\ref{listing:spatial_relations} depicts how the ChangeFormer detections are integrated into the HPLP as a new spatial relation \textit{p::change(x)}.
Here, each output probability is analogous to the demonstration of the \textit{over(x, t)} relation, written as the conditional probability of the HPLP fact.
In contrast to prior work, the relation is unary and not connected to a specific type due to the less specific semantics of the ChangeFormer.

\input{listings/uam_model}
\input{figures/pml_matrix}
\input{figures/scenarios}

\subsection{Reasoning on Neuro-Symbolic Navigation Spaces}

In the previous two sections, we demonstrated the creation of neuro-symbolic spatial relations based on the statistical evaluation of geographic data (\textit{distance(x, t)}, \textit{over(x, t)}) and neural perception (\textit{change(x)}).
Now, we demonstrate how this data is integrated for various locations and application scenarios into HPLPs and present possible outcomes of employing ProMis.

Consider Listing~\ref{listing:uam_model} as a simple encoding of flight regulations.
While this model has been handcrafted to illustrate the effect of modeling AAM restrictions with ProMis, we later show its creation through prompting an LLM (see Section~\ref{sec:llm}).
The model consists of four sections.

First, the UAS's properties are encoded as estimates of its mass, battery charge, and drain per distance flown. 
Second, definitions of what circumstances will constitute visual line-of-sight (VLOS) and whether the UAS has enough charge left to return to the operator are given.
Third, we define a set of permissions granted for the operation of the UAV using the \textit{distance(x, t)} and \textit{over(x, t)} relations estimated before.
Finally, the \textit{landscape(X)} is specified based on the stated clauses, combining the concepts introduced so far.

ProMis facilitates interpretable and adaptable mission design by allowing easy reconfiguration of the model, meaning rules may be rewritten and constraint boundaries altered through human intervention without necessitating learning or a deeper mathematical understanding.
Figure~\ref{fig:pml_matrix} highlights this principle by visualizing parameters of some of the spatial relations computed for the first scenario shown in Figure~\ref{fig:scenarios}.

A single model can be applied to many environments without change, e.g., if the same public regulations or operator preferences apply to each.
Here, we consider four settings shown in Figure~\ref{fig:scenarios}, covering scenarios from leisure UAS activities in a city park, flights across a bay area in the Baltic Sea, applications in a dense urban canyon, and inspection flights over a railroad network.

One can see how, with a single model of flight regulations, ProMis can generate appropriate PMLs without requiring local adaptations.
For example, given a public encoding of a country's aviation laws, operators can confidently employ ProMis independently of the application's location.

In each scenario, the operator's location was placed at the origin of the coordinate frame.
We apply the same parameters, e.g., resolution and error model, as in Section~\ref{sec:distance_over_exp}.

\input{figures/individual_times}
\input{figures/error}

\subsection{Runtime Analysis and Parallelization}

Understanding ProMis' applicability to real-world tasks requires understanding its runtime characteristics.
There are three key considerations: the number of sample points, the number of cores, and the batch size.
All measurements have been obtained on an 8-core Intel i7 9700K.

ProMis generates each sample of the PML in an independent and identically distributed manner.
Thus, we can parallelize the inference of the PML to reduce computation time while retaining exact inference (see Figure~\ref{fig:runtime-parallel}).

ProMis inference time is linear with the number of samples. 
Hence, when sampling a 2D grid with growing resolution, we observe a quadratic increase in runtime (see Figure~\ref{fig:runtime-resolution}).
Therefore, leveraging a smarter sampling technique can be important for an efficient application of ProMis. 

In both experiments, one can see that there is an optimal choice of batch size with diminishing impact at higher resolutions and when employing more cores.

Finally, we evaluate the runtime requirements of ProMis split into the parameter estimation and inference (see Figure~\ref{fig:runtime-steps}.
One can see that both contribute nearly equally to the total runtime.

\subsection{Adaptive Sampling of Probabilistic Mission Landscapes}

\input{figures/llm}

Sampling an accurate PML is crucial for a mission's success and safety. 
However, the ideal number and distribution of samples heavily dependon the scenario.
Thus, we analyze sample efficiency and the loss of accuracy compared to a ground-truth reference PML computed in a high resolution ($200 \times 200$ samples) in the previously shown scenarios of Figure~\ref{fig:scenarios}.
We do so by running ProMis with an initial $10 \times 10$ PML and comparing three incremental sampling approaches based on the Mean Absolute Error (MAE).

(i) A naive, grid-wise sampling with increasing resolution, (ii) local entropy as acquisition function, determining locations of possibly high information gain, and (iii) guiding sampling via a Gaussian Process (GP), sampling where the covariance of the GP is highest.
For comparison, each PML is upscaled using either linear or nearest-neighbor interpolation, and the MAE compared to the reference is obtained.
While Figure~\ref{fig:adaptive_sampling} showcases this process for applying local entropy as an acquisition function, Figure~\ref{fig:error} shows the MAEs for each scenario individually as well as the relative improvement in MAE of the adaptive procedures compared to the naive grid.

\subsection{Probabilistic Mission Design with Large Language Models}
\label{sec:llm}

Because ProMis allows us to limit the operations of an agent via first-order logic, we can leverage the fact that Large Language Models (LLM) can generate program code from natural language descriptions.
We employ DeepSeek\footnote{https://chat.deepseek.com, accessed January 31, 2025} as an LLM to demonstrate this approach.
To do so, we must tell the LLM what a ProMis program is and describe its elements.
For this, we pass Listing~\ref{listing:uam_model} together with the request of translating the following prompts into equivalent ProMis code, underlining the need for the \textit{landscape} clause to be included and to use the \textit{distance}, \textit{over}, and \textit{change} spatial relations as exemplified.

Given the mission description of an operator who is unable to write the first-order logic required by ProMis themselves, the LLM generates the requested encoding.
Figure~\ref{fig:llm} shows an example prompt, the LLM's response, including the generated program, and the computed PML in New York's Central Park.

While creating a fine-tuned LLM may be desirable to avoid semantic errors during generation, an advantage of ProMis employing logic programs is that any syntax errors will be caught during compilation, so they cannot propagate into a PML.
Furthermore, errors reported by the solver may be returned to the LLM to improve the code.

%% file: listings/uam_model.tex
\begin{listing}
    \centering
    \begin{minted}
    [
        frame=none,
        autogobble,
        fontsize=\footnotesize,
    ]{prolog}
    % UAV properties
    initial_charge ~ normal(90, 5).
    charge_cost ~ normal(-0.1, 0.2).
    weight ~ normal(2.0, 0.1).

    % Visual line of sight
    1/10::fog; 9/10::clear.
    vlos(X) :- 
        fog, distance(X, operator) < 50;
        clear, distance(X, operator) < 100;
        clear, over(X, bay), distance(X, operator) < 400.

    % Sufficient charge to return to operator
    can_return(X) :-
        B is initial_charge, O is charge_cost,
        D is distance(X, operator), 0 < B + (2 * O * D).

    % Permits related to local features
    permits(X) :- 
        distance(X, service) < 15; 
        distance(X, primary) < 15;
        distance(X, secondary) < 10; 
        distance(X, tertiary) < 5;
        distance(X, crossing) < 5; 
        distance(X, rail) < 5;
        over(X, park).

    % Definition of a valid mission
    landscape(X) :- 
        vlos(X), weight < 25, can_return(X); 
        permits(X), can_return(X).
    \end{minted}
    \caption{
        Operator-written model for urban AAM scenario.
        The resulting PMLs are shown in Figure~\ref{fig:scenarios}.
    }
    \label{listing:uam_model}
\end{listing}

%% file: figures/pml_matrix.tex
\begin{figure*}
    \centering
    \begin{subfigure}{0.27\textwidth}
        \includegraphics[width=\textwidth]{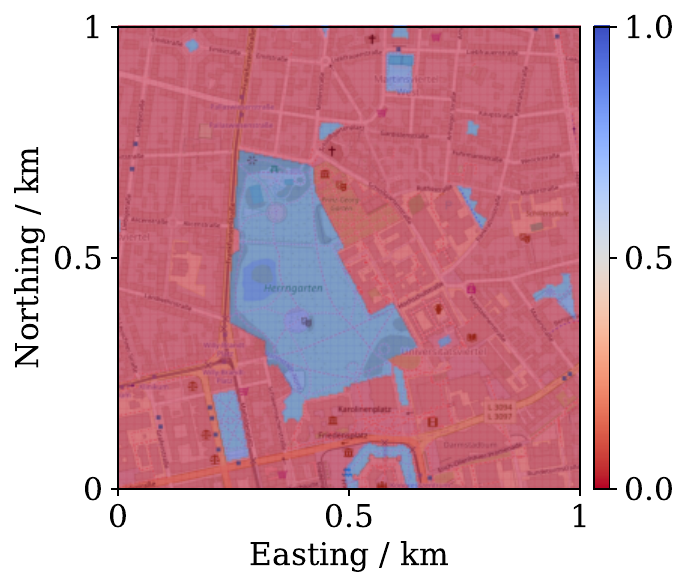}
        \caption{$P(over(X, park))$}
    \end{subfigure} 
    \begin{subfigure}{0.235\textwidth}
        \includegraphics[width=\textwidth]{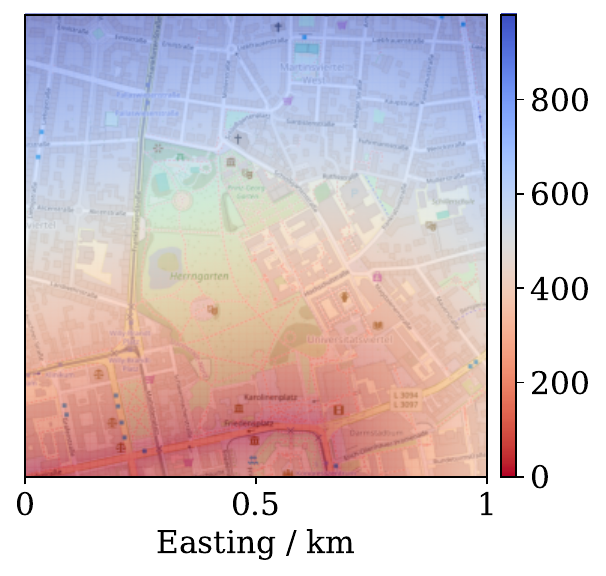}
        \caption{$\mu$ of $dist(X, primary)$}
    \end{subfigure}
    \vspace{0.35em}
    \begin{subfigure}{0.235\textwidth}
        \includegraphics[width=\textwidth]{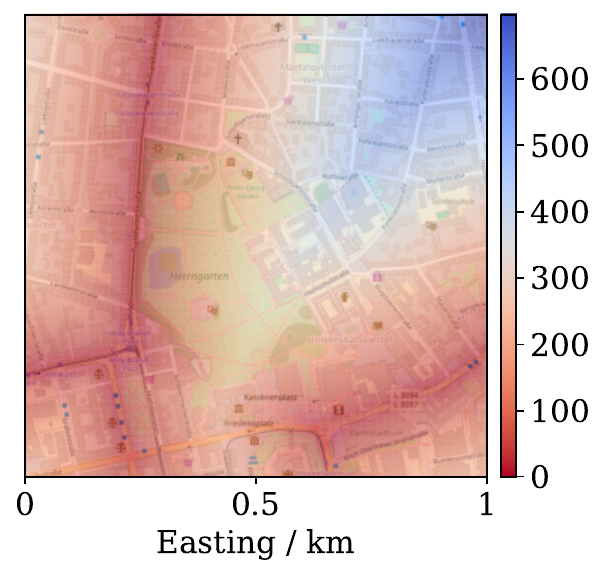}
        \caption{$\mu$ of $dist(X, secondary)$}
    \end{subfigure} 
    \begin{subfigure}{0.235\textwidth}
        \includegraphics[width=\textwidth]{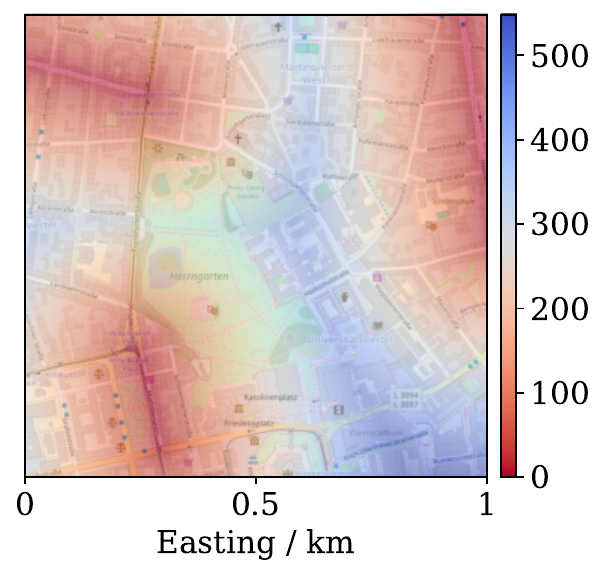}
        \caption{$\mu$ of $dist(X, tertiary)$}
    \end{subfigure} 
    \caption{
        \textbf{Adaptable mission design with ProMis:} 
        Spatial relations express environment features in a hybrid probabilistic fashion and provide a vocabulary or encoding flight regulations.
        While some express categorical information, such as probabilistic occupancy (a), others allow adapting flight rules with variable boundaries relative to the sets of underlying geographic features (b-d).
        Hence, while Figure~\ref{fig:scenarios-darmstadt} shows a PML in this environment based on Listing~\ref{listing:uam_model}, one can create arbitrary desired combinations of rules from this vocabulary.
    }
    \label{fig:pml_matrix}
\end{figure*}

%% file: figures/scenarios.tex
\begin{figure*}
    \centering
    \begin{subfigure}{0.27\textwidth}
        \includegraphics[width=\textwidth]{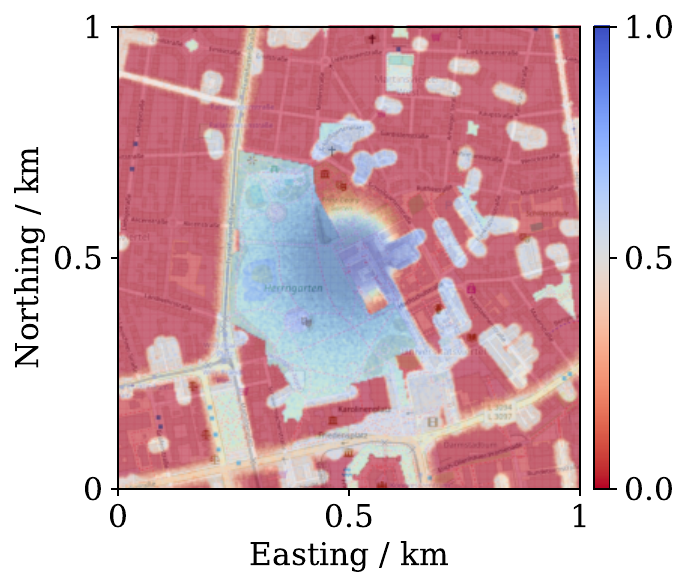}
        \caption{Darmstadt University District}
        \label{fig:scenarios-darmstadt}
    \end{subfigure}
    \begin{subfigure}{0.235\textwidth}
        \includegraphics[width=\textwidth]{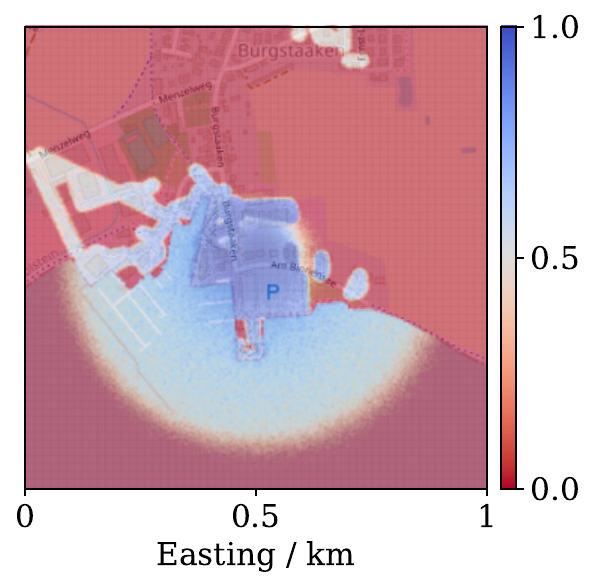}
        \caption{Fehmarn Bay Area}
    \end{subfigure}
    \begin{subfigure}{0.235\textwidth}
        \includegraphics[width=\textwidth]{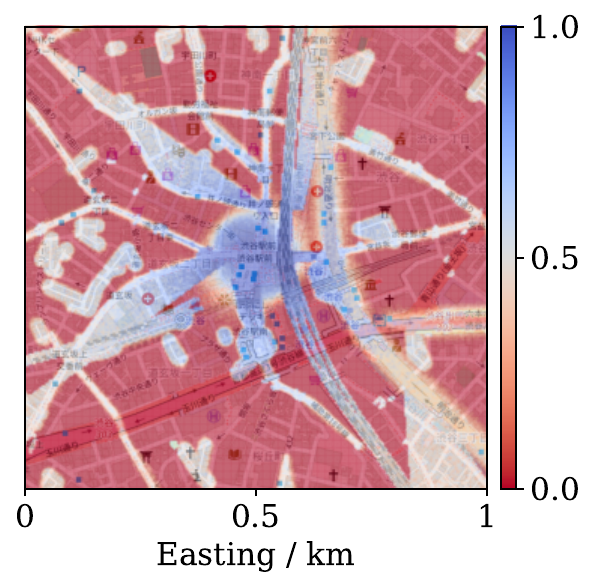}
        \caption{Shibuya Crossing}
    \end{subfigure}
    \begin{subfigure}{0.235\textwidth}
        \includegraphics[width=\textwidth]{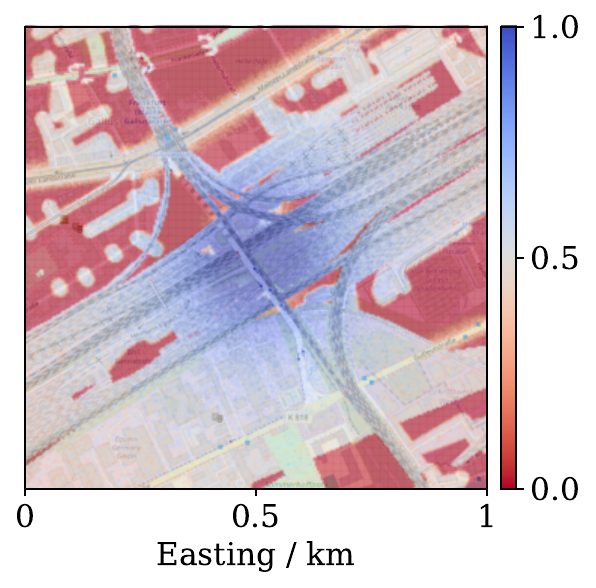}
        \caption{Frankfurt Central Station}
    \end{subfigure}
    \caption{
        \textbf{ProMis generalizes to a variety of scenarios:}
        We show the result of utilizing ProMis in (a) an urban city park with adjacent major road sections, (b) a bay area with increased visual-line-of-sight over open waters and accessible service roads, and (c) a major junction with pedestrian crossings where a green signal allows crossing one from side to the other.
        Finally, (d) shows a PML across passenger and cargo rails near a train station.
    }
    \label{fig:scenarios}
\end{figure*}

%% file: figures/individual_times.tex
\begin{figure*}
    \centering
    \begin{subfigure}{0.32\textwidth}
        \includegraphics[width=\textwidth]{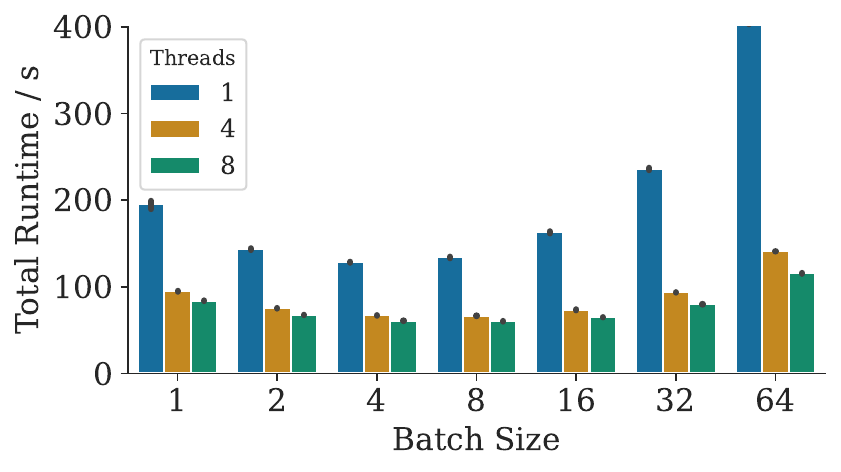}
        \caption{Varying Parallelization}
        \label{fig:runtime-parallel}
    \end{subfigure}
    \begin{subfigure}{0.32\textwidth}
        \includegraphics[width=\textwidth]{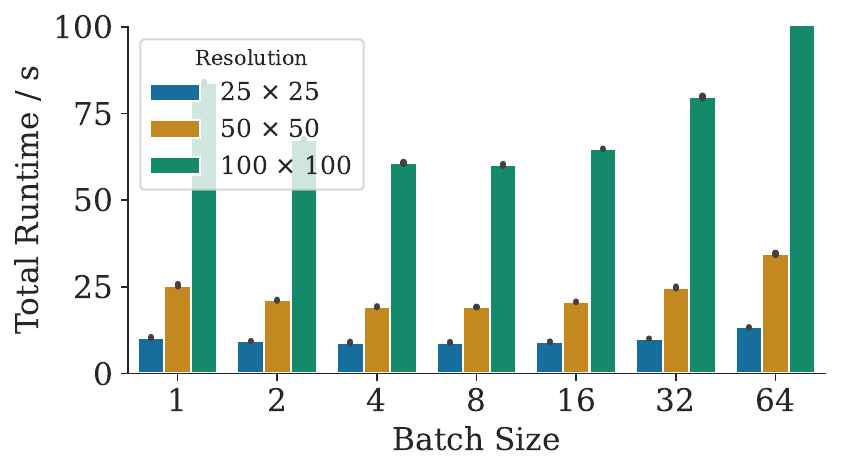}
        \caption{Varying Resolution}    
        \label{fig:runtime-resolution}
    \end{subfigure}
    \begin{subfigure}{0.32\textwidth}
        \includegraphics[width=\textwidth]{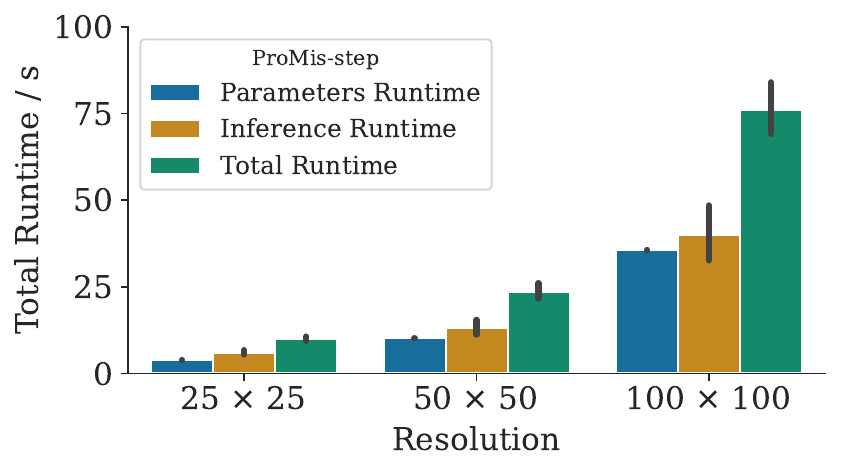}
        \caption{Runtime per ProMis-step}
        \label{fig:runtime-steps}
    \end{subfigure}
    \caption{
        \textbf{ProMis runtime analysis:}
        We evaluate the runtime requirements of ProMis for a fixed $100 \times 100$ resolution and a varying number of threads (a), a fixed number of $8$ threads on a varying resolution (b), and the time required for parameter estimation, inference and overall runtime at a fixed number of threads and batch size (c).
    }
    \label{fig:runtime}
\end{figure*}

%% file: figures/error.tex
\begin{figure*}
    \centering
    \includegraphics[width=0.8\textwidth]{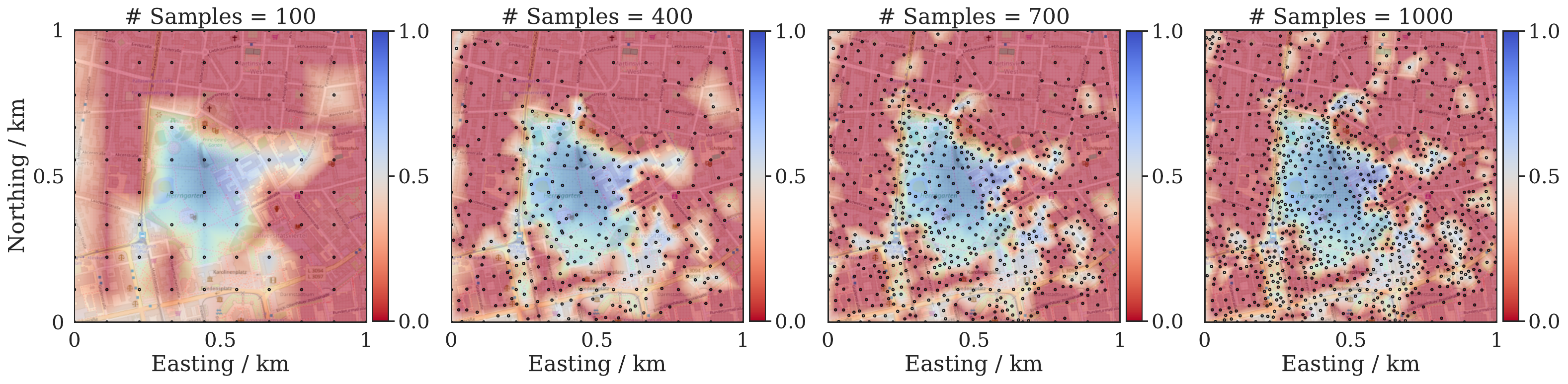}
    \caption{
        \textbf{Adaptive sampling in ProMis:}
        Here, we demonstrate the effect of applying an acquisition function such as local entropy for adaptively sampling (black dots) in ProMis, leading to higher sample efficiency compared to a naive rasterization.
    }
    \label{fig:adaptive_sampling}
\end{figure*}

\begin{figure*}
    \centering
    \begin{subfigure}{0.23\textwidth}
        \includegraphics[trim=0 0 0 0, width=\textwidth]{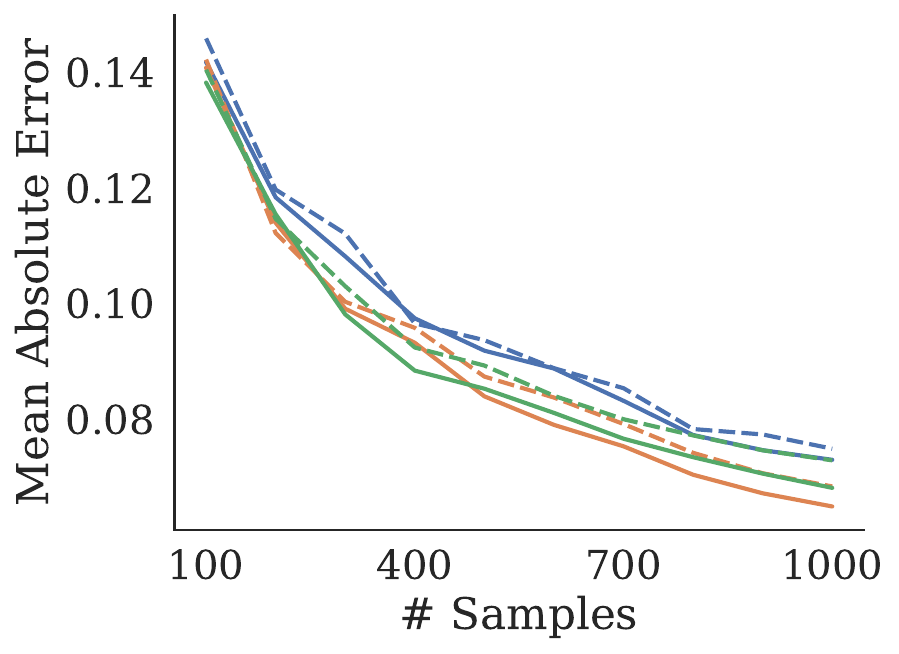}
    \end{subfigure}
    \hfill
    \begin{subfigure}{0.225\textwidth}
        \includegraphics[width=\textwidth]{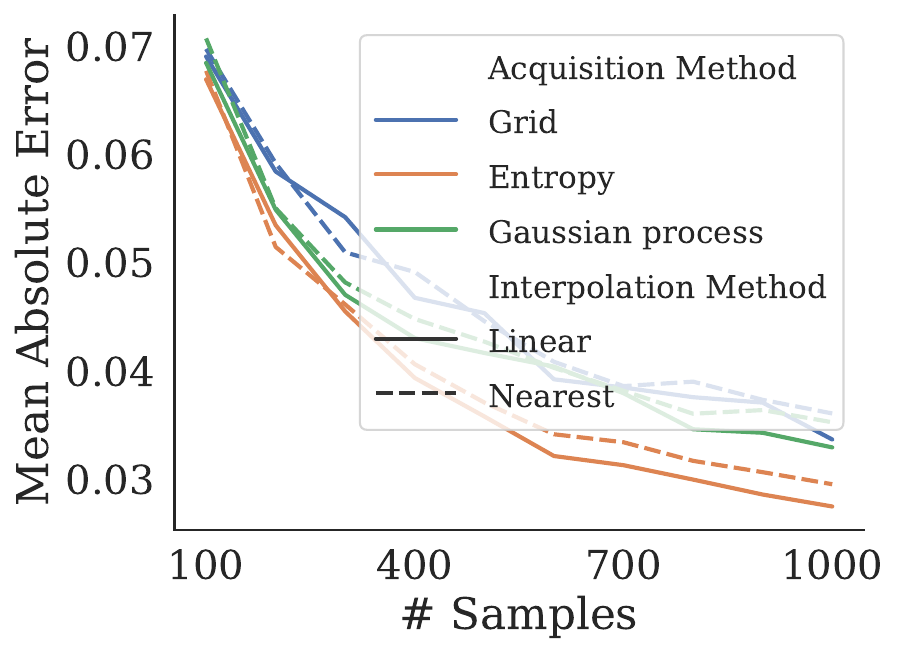}
    \end{subfigure}
    \hfill
    \begin{subfigure}{0.225\textwidth}
        \includegraphics[width=\textwidth]{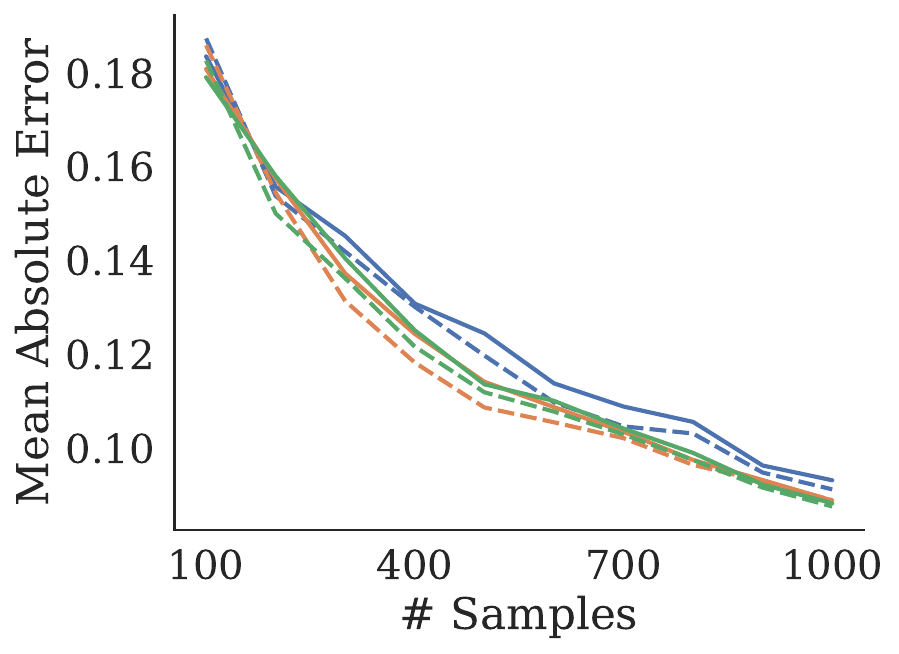}
    \end{subfigure}
    \hfill
    \begin{subfigure}{0.225\textwidth}
        \includegraphics[width=\textwidth]{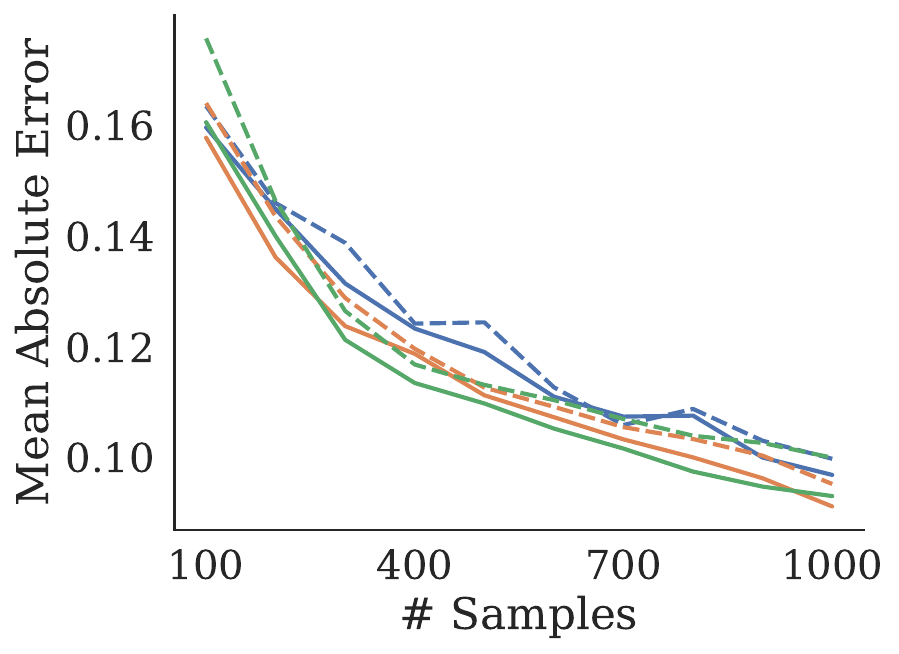}
    \end{subfigure}\\
    \begin{subfigure}{0.23\textwidth}
        \includegraphics[trim=0 0 0 0, width=\textwidth]{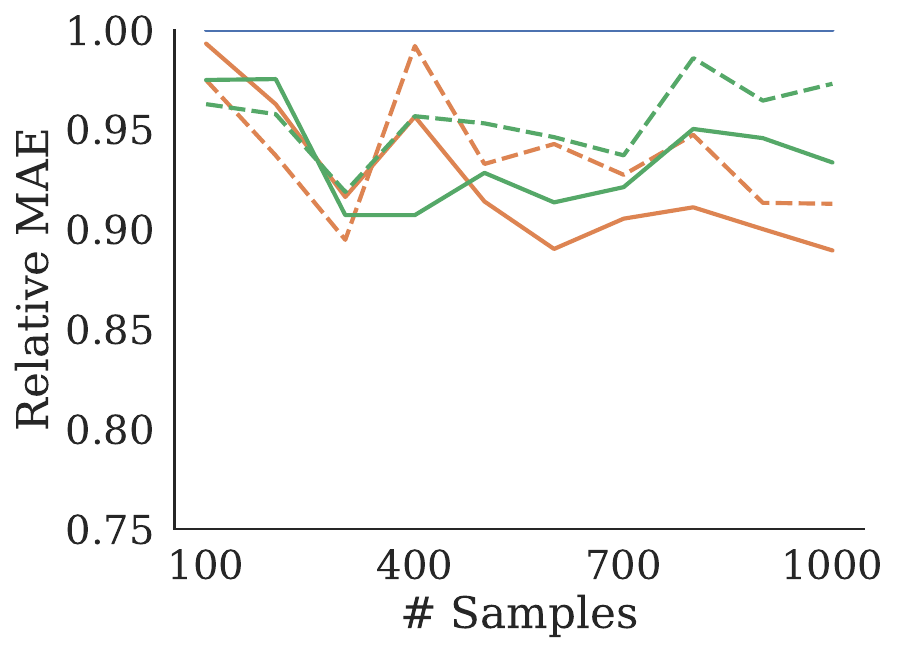}
        \caption{Darmstadt Interpolation}
    \end{subfigure}
    \hfill
    \begin{subfigure}{0.225\textwidth}
        \includegraphics[width=\textwidth]{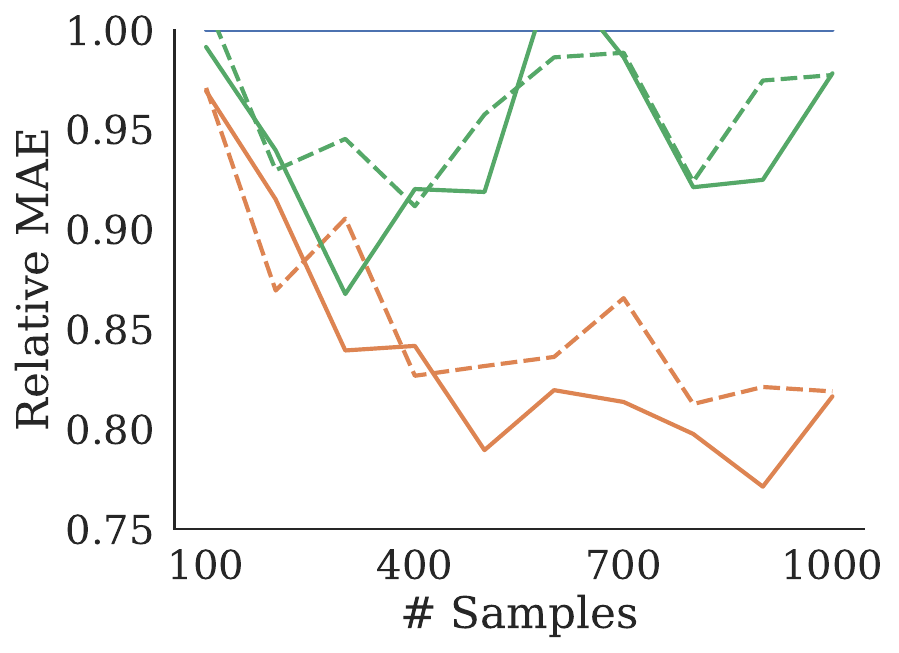}
        \caption{Fehmarn Interpolation}    
    \end{subfigure}
    \hfill
    \begin{subfigure}{0.225\textwidth}
        \includegraphics[width=\textwidth]{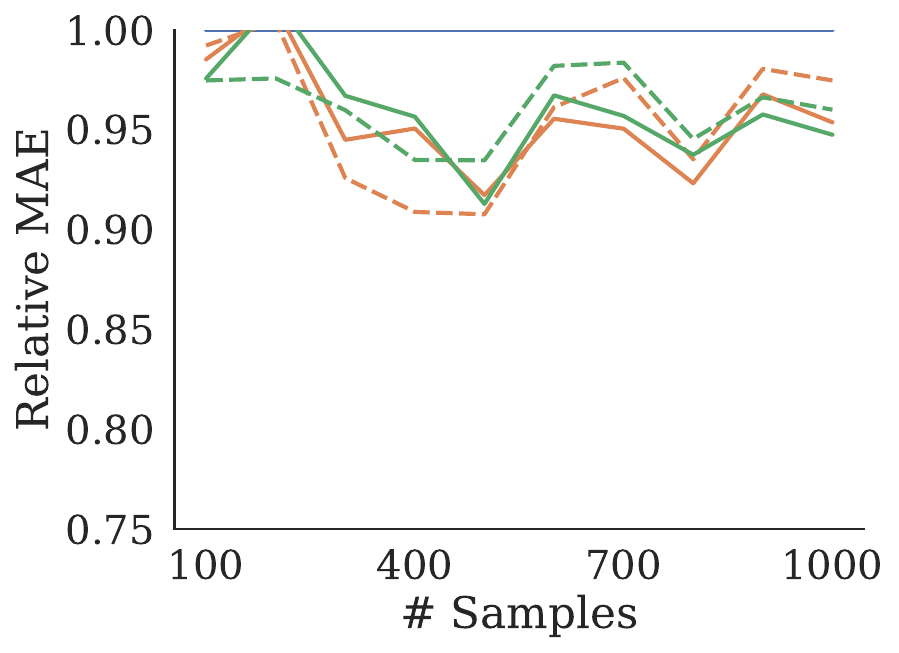}
        \caption{Shibuya Interpolation}
    \end{subfigure}
    \hfill
    \begin{subfigure}{0.225\textwidth}
        \includegraphics[width=\textwidth]{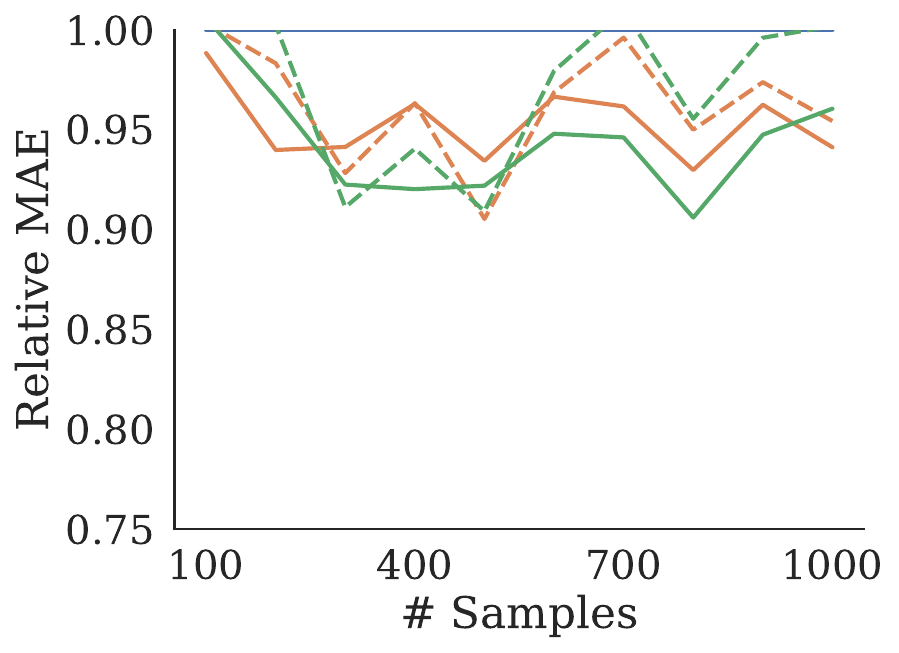}
        \caption{Frankfurt Interpolation}
    \end{subfigure}
    \caption{
        \textbf{Comparison study:}
        We consider grid-wise rasterization, entropy, and Gaussian Process based sampling combined with linear and nearest-neighbour interpolation.
        The first row shows the MAE for each method, while the second visualizes the advantage of adaptive sampling relative to grid-wise sampling when compared to a $200 \times 200$ ground-truth PML.
    }
    \label{fig:error}
\end{figure*}

%% file: figures/llm.tex
\begin{figure}
    \centering
    \includegraphics[width=0.8\linewidth]{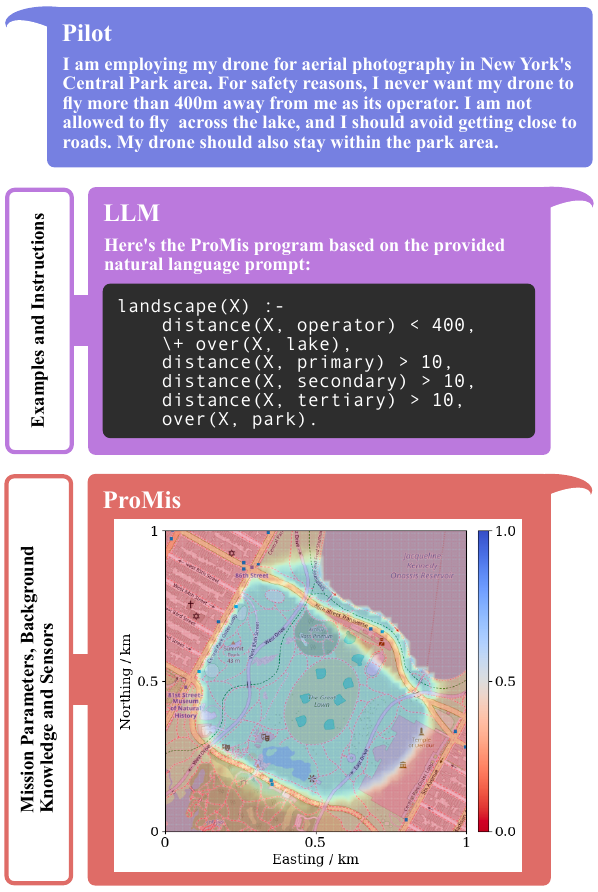}
    \caption{
        \textbf{Large Language Model integration:}
        LLMs can encode a natural language description of flight requirements by providing them with examples and instructions beforehand.
        Hence, LLMs pave the way for safe UAS applications for non-experts and complex or changing regulations.
    }
    \label{fig:llm}
\end{figure}%

%% file: content/5_conclusion.tex
\section{Conclusion}
\label{sec:conclusion}

We have presented Probabilistic Mission Design (ProMis) for neuro-symbolic transportation systems, such as advanced Unmanned Aerial Vehicles, based on Hybrid Probabilistic Logic Programs.
ProMis facilitates adaptable and interpretable mission design by encoding multi-modal input data into spatial relations for encoding flight regulations.
Probabilistic Mission Landscapes (PML) quantify how likely public, operator, or LLM-encoded rules are satisfied under uncertainty.
Hence, they facilitate advanced mission design tasks such as granting clearance, explaining issues, or optimizing missions~\cite{kohaut2024ceo}.

ProMis' application in dynamic scenarios is currently limited by its runtime requirements for parameter estimation and inference, necessitating more efficient sampling and inference techniques in the future.
While ProMis is well equipped to deal with sensor noise and varying map quality, its probabilistic semantics are dependent on scenarios where uncertainty data, e.g., in the map, is provided.
Furthermore, the natural language interface to ProMis through LLMs may benefit significantly from developing fine-tuned expert models.
Along this line, one may consider more interactive settings, where the operator and LLM iterate on a landscape definition.

%% file: content/6_acknowledgment.tex
\section*{Acknowledgments}

The Eindhoven University of Technology authors received support from their Department of Mathematics and Computer Science and the Eindhoven Artificial Intelligence Systems Institute.
Map data \copyright~OpenStreetMap contributors, licensed under the ODbL and available from \url{www.openstreetmap.org}.